\PassOptionsToPackage{table}{xcolor}
\documentclass[sigconf]{acmart}

\setcopyright{none}
\renewcommand\footnotetextcopyrightpermission[1]{}

\AtBeginDocument{%
  }

\usepackage{graphicx}
\usepackage{booktabs}
\usepackage{multirow}
\usepackage{makecell}
\usepackage{amsmath}

\copyrightyear{2026}
\acmYear{2026}
\setcopyright{cc}
\setcctype{by}
\acmConference[GreenMM '26]{Efficient Systems and AI for the Planet}{November 10--14, 2026}{Rio de Janeiro, Brazil}
\acmBooktitle{Efficient Systems and AI for the Planet (GreenMM '26), November 10--14, 2026, Rio de Janeiro, Brazil}
\acmDOI{10.1145/3840750.3841444}
\acmISBN{979-8-4007-2936-2/2026/11}

\begin{document}

\title{Geo-LoRA: Geometry-Aware Subspace Evolution for Low-Rank Adaptation in
Continual Learning}

\author{Yibo Feng}
\affiliation{%
  \institution{University of Electronic Science and Technology of China}
  \city{Chengdu}
  \country{China}
}
\email{2022090917012@std.uestc.edu.cn}


\renewcommand{\shortauthors}{Yibo Feng}

\begin{abstract}
 Rehearsal-free class-incremental learning (CIL) with LoRA adapters remains challenging because the low-rank subspaces updated across tasks evolve without geometric control, causing unstable shared representations and repetitive collapse of task-specific updates into previously occupied directions. We introduce Geo-LoRA, a geometry-aware framework that explicitly regulates how low-rank subspaces—shared and task-specific—evolve during continual learning.
For the shared branch, Subspace Projection Preservation (SPP) constrains consecutive updates to follow smooth trajectories on the Grassmann manifold, and Adaptive Core–Slack Alignment (ACSA) decomposes transitions into principal and residual components, aligning the former while modulating the latter to balance stability and plasticity. For the task-specific branch, Median-Calibrated Block Overlap (MCBO) imposes a statistical constraint via normalized projection overlap, penalizing excessive reuse to mitigate subspace crowding.
These constraints jointly regulate the evolution of all LoRA subspaces across layers and tasks without introducing additional adapter types beyond standard LoRA. Geo-LoRA provides a principled geometric formulation for continual low-rank adaptation and consistently achieves state-of-the-art performance across multiple benchmark datasets and different task lengths.
\end{abstract}





\maketitle
\section{Introduction}
\label{sec:intro}

Class-incremental learning (CIL)\cite{zhou2024class, he2024gradient, li2024fcs, li2017learning} requires models to learn new classes over time without forgetting previously acquired knowledge. With the rise of pre-trained vision transformers (ViTs), recent work increasingly adopts parameter-efficient fine-tuning (PEFT)\cite{gao2024beyond, araujo2024learning, zhang2023adapter} to enable exemplar-free CIL, where the backbone remains frozen and lightweight modules handle continual adaptation. Among PEFT techniques, Low-Rank Adaptation (LoRA)\cite{hu2022lora} is particularly attractive due to its efficiency and its interpretation as learning task-dependent low-dimensional feature subspaces.

However, although existing LoRA-based CIL methods\cite{zhu2025bilora, wu2025sdlora, yu2025fm} have highlighted interference among low-rank updates, they still provide limited explicit control over the geometric evolution of these low-rank subspaces.
Most approaches attach independent task-specific LoRA modules for each task, while recent dual-branch designs\cite{he2025cl} introduce shared adapters across tasks. In both cases, low-rank directions learned at different stages are optimized mostly in isolation: shared subspaces may drift across tasks, and deep task-specific updates may repeatedly occupy previously used directions. These effects lead to misaligned shared representations and subspace crowding in layers, undermining the stability\cite{mccloskey1989catastrophic}–plasticity\cite{dohare2024loss, wang2024improving} trade-off.

These observations motivate a different perspective. Instead of viewing continual LoRA tuning as merely adding or freezing adapters, we formulate it as regulating the evolution of low-rank subspaces across tasks. This geometric viewpoint forms the basis of Geo-LoRA. Rather than modifying the LoRA architecture itself, Geo-LoRA introduces projection-based geometric objectives that guide low-rank updates to evolve within a stable yet flexible envelope.

\begin{figure}[t]
  \centering
  \includegraphics[width=0.42\textwidth]{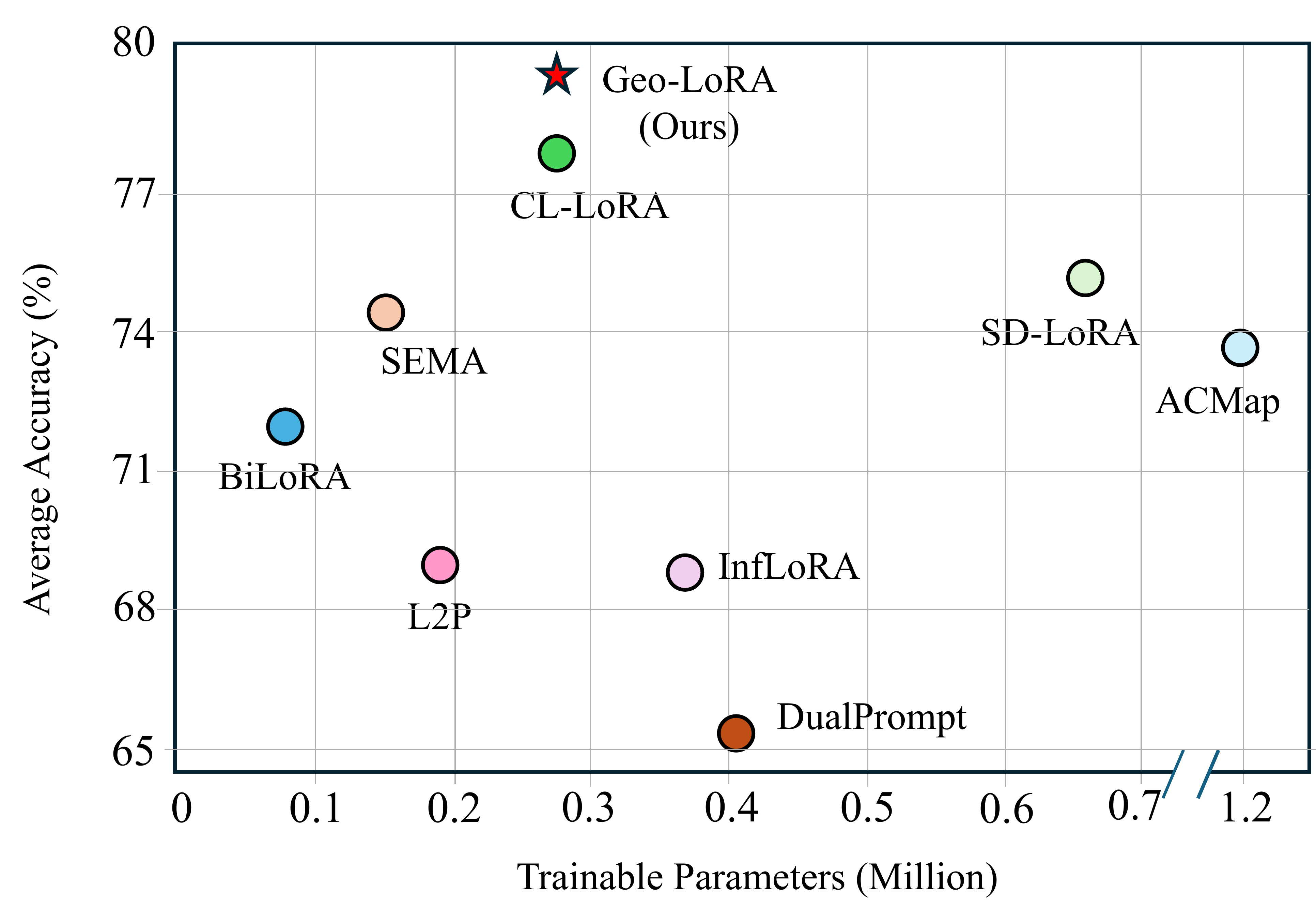}
  \caption{Comparison of different PTM-based CIL methods on ImageNet-R dataset (20 tasks). The X-axis stands for the number of trainable parameters, and the Y-axis represents the average accuracy. For a fair comparison, all methods are based on pre-trained ViT-B/16-IN1K.}
  \label{fig:shuffle}
\end{figure}

To enable such geometric control, we adopt the Dual-LoRA architecture from CL-LoRA\cite{he2025cl} as a structural backbone. This design separates continual adaptation into two regimes: shallow layers maintain a shared low-rank subspace that accumulates transferable structure, while deeper layers host task-specific LoRA updates for semantic specialization. This decomposition allows different geometric constraints to regulate shared and task-specific subspaces. Importantly, our goal is not to propose a new LoRA architecture, but to introduce geometry-aware objectives that stabilize and structure the learning dynamics within this framework.

Within this hierarchy we design three complementary objectives. Subspace Projection Preservation (SPP) constrains consecutive shared updates using Grassmann-consistent projection distance to reduce cross-task drift. Adaptive Core–Slack Alignment (ACSA) aligns principal shared directions while adaptively moderating residual directions to balance stability and plasticity. Median-Calibrated Block Overlap (MCBO) regulates deep-layer updates through normalized projection overlap, penalizing only statistically excessive reuse of historical directions. Together these components form a multi-level geometric regularization framework that controls subspace evolution without introducing additional adapter types and with modest training overhead.

As illustrated in Figure~\ref{fig:shuffle}, Geo-LoRA achieves state-of-the-art (SOTA) performance while maintaining a small number of trainable parameters.

Our key contributions are summarized as follows:

(1) We provide a geometric subspace-evolution perspective on LoRA-based CIL, offering a principled view of continual low-rank adaptation.

(2) We propose Geo-LoRA, integrating SPP, ACSA, and MCBO into the Dual-LoRA architecture to regularize shared and task-specific subspaces under a unified projection-based principle.

(3) We achieve SOTA results across multiple datasets and tasks, demonstrating that geometry-aware subspace regulation provides an effective and scalable paradigm for exemplar-free continual learning.

\section{Related work}
\subsection{Parameter-Efficient Fine-Tuning}
Class-incremental learning (CIL) aims to acquire new classes over time without forgetting previously learned ones.
Recent work increasingly leverages large pre-trained vision transformers (ViTs), whose strong general-purpose representations reduce the need to store exemplars.
Parameter-efficient fine-tuning (PEFT) approaches, such as adapters\cite{sun2025mos, qi2025adaptive, fukuda2025adapter, liu2023class, panos2023first, liu2024cala}, prompt tuning\cite{xu2025c, huang2024learning, jiang2025revisiting, liu2024few, huang2024learning, akmel2024few, huang2024class}, and LoRA—enable continual adaptation while keeping the backbone frozen, thereby limiting full-parameter interference across tasks.
However, many PEFT-based CIL methods still rely on task-specific modules that grow linearly with the number of tasks, leading to parameter overhead and fragmented cross-task representations.
A few recent studies explore shared components across tasks, but they typically regularize parameters\cite{he2025cl} or optimize prediction objectives at the module level\cite{wang2022dualprompt}, rather than explicitly modeling how low-rank feature subspaces evolve over time.
Moreover, recent LoRA-based CIL methods, including InfLoRA\cite{liang2024inflora} and related approaches\cite{wang2023orthogonal, qiao2024learn, bhat2025parameter, he2025cl}, have already highlighted the importance of interference among low-rank updates.
Our work builds on this line of research, but differs in focusing explicitly on the geometric evolution of both shared and task-specific LoRA subspaces across tasks.

\subsection{Geometric Evolution in LoRA}
LoRA adapts frozen weights through low-rank matrices that can be interpreted as learning new directions in a task-dependent feature subspace\cite{aghajanyan2021intrinsic, he2021towards}.
In multi-task or continual scenarios, these low-dimensional directions may rotate, collapse, drift, or interfere with one another, leading to misaligned representations and degraded transfer\cite{wang2023orthogonal, qiao2024learn, bhat2025parameter}.
Prior work has attempted to alleviate such effects using orthogonality, sparsity, adaptive rank allocation, or shared/task-specific branch designs, but these methods mainly regulate parameters, overlap, or module composition rather than directly controlling the trajectories of the underlying low-rank subspaces\cite{he2025cl}.
Subspace-based formulations have also appeared in domain adaptation\cite{chen2021representation, fernando2013unsupervised, zhang2019guide}, yet they typically operate on feature covariance or classifier weights instead of LoRA-induced subspaces under continual task sequences.
Moreover, existing methods rarely distinguish between principal directions that should remain stable and residual directions that can support plasticity, nor do they address the block-wise imbalance caused when deep adapters repeatedly reuse historical subspaces.
Geo-LoRA addresses these gaps through three projection-based mechanisms—SPP, ACSA, and MCBO—that explicitly regulate shared drift, core-versus-slack evolution, and excessive historical reuse, providing a unified geometric perspective on continual low-rank adaptation.

\section{Preliminaries}

In rehearsal-free class-incremental learning (CIL), a model learns a sequence of tasks with disjoint label spaces without storing past samples. A pre-trained ViT $f_\theta$ is frozen, and lightweight adapters are inserted into Transformer blocks:
\begin{equation}
z = f_\theta(x) + f_{\mathrm{adapt}}(x).
\end{equation}
LoRA implements each adapter via a rank-$r$ update $AB$ to a frozen weight $W$:
\begin{equation}
z = Wx + ABx,\quad r \ll \min(d,k).
\end{equation}
Most CIL methods allocate task-specific LoRA modules $(A_t,B_t)$ for every task, causing linear parameter growth and limited knowledge sharing.

Geo-LoRA follows the Dual-LoRA architecture introduced in prior work, but reinterprets it through a geometric viewpoint. The first $l$ blocks contain a single shared LoRA module used across all tasks, while deeper blocks host task-specific LoRA modules. Rather than treating these as independent parameter augmentations, Geo-LoRA regards both branches as evolving low-rank subspaces: the shallow shared subspace should evolve smoothly across tasks, whereas deep task-specific subspaces should either expand or reuse prior directions in a controlled way.

Formally, the feature update for task $t$ at block $i$ is
\begin{equation}
z_i^t = f_\theta^i(z_{i-1}^t) +
\begin{cases}
A_s^i B_s^i z_{i-1}^t, & i \le l, \\
A_t^i B_t^i z_{i-1}^t, & i > l.
\end{cases}
\end{equation}

This geometric reinterpretation provides the foundation for the three regularizers introduced next—SPP, ACSA, and MCBO—which collectively regulate how shared and task-specific LoRA subspaces evolve throughout continual learning.

\section{Method}

\begin{figure*}[!ht]
    \begin{center}
	\includegraphics[width=1.0\linewidth]{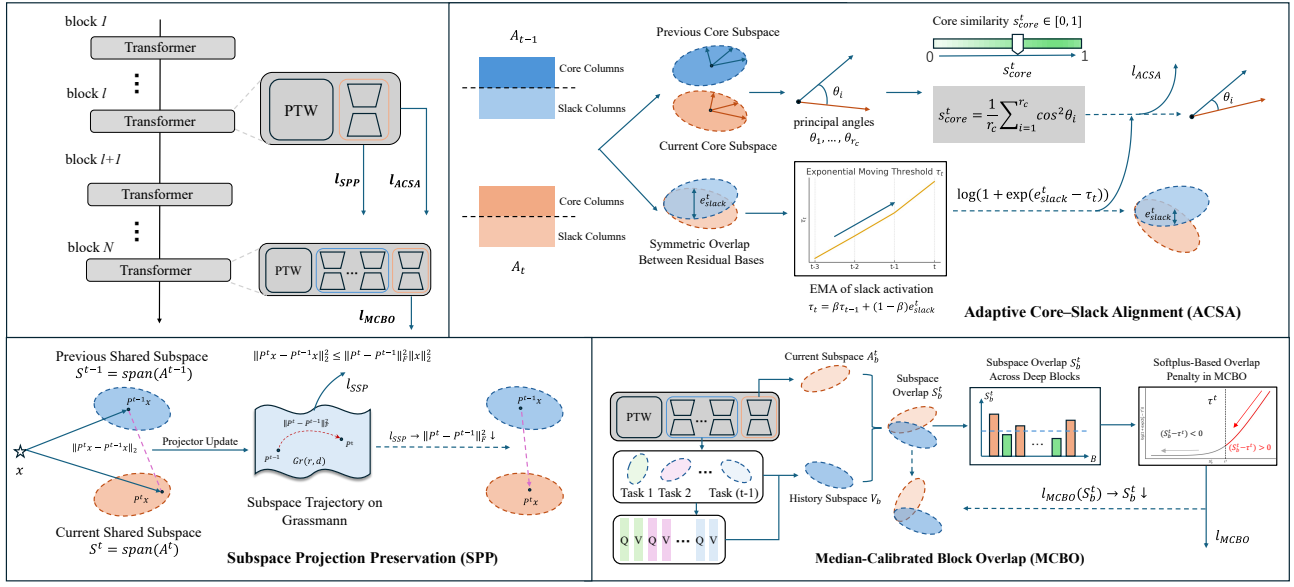}
	\caption{Overview of Geo-LoRA.
Geo-LoRA regulates the geometric evolution of LoRA-induced subspaces through three lightweight constraints.
(1) \textbf{SPP} enforces smooth transitions of the shared low-rank subspace by minimizing Grassmannian projector drift.
(2) \textbf{ACSA} decomposes subspace changes into core and slack components, preserving aligned directions while adaptively controlling residual activation.
(3) \textbf{MCBO} suppresses excessive reuse of historical directions in deep task-specific blocks via median-calibrated normalized overlaps.
Together, these objectives coordinate stable yet flexible subspace evolution across tasks without increasing inference cost.}
	\label{fig:method}
    \end{center}
\end{figure*}

In this section, we present Geo-LoRA, a unified geometric framework for continual low-rank adaptation, and Figure~\ref{fig:method} summarizes its workflow.
Instead of introducing separate LoRA modules for each task, Geo-LoRA explicitly constrains how LoRA-induced low-rank subspaces evolve across tasks.
We adopt a dual-LoRA architecture with (i) a shallow \emph{shared} branch for cross-task knowledge preservation and (ii) a deep \emph{task-specific} branch for task-level specialization.
Both branches are regularized by lightweight geometric constraints that balance stability and plasticity without adding inference-time overhead.

\subsection{Subspace Projection Preservation}

In Geo-LoRA, shallow LoRA modules are shared across tasks and continually refined, accumulating transferable structure in early Transformer layers.
However, continual updates can drift the underlying \emph{subspace}, so an input feature previously represented in $S_{t-1}$ may be mapped to a rotated $S_t$.
Since class prototypes from earlier tasks are fixed under $S_{t-1}$, such rotations induce feature--prototype misalignment and exacerbate forgetting.

To counter this, we regularize the evolution of the shallow shared subspaces via Subspace Projection Preservation (SPP).
After finishing task $t{-}1$, we concatenate the up-projection matrices of all shallow shared LoRA modules to form
$A_{t-1}\!\in\!\mathbb{R}^{d\times q_{t-1}}$, and analogously obtain $A_t\!\in\!\mathbb{R}^{d\times q_t}$ after finishing task $t$.
Here $q_t$ denotes the total concatenated width (i.e., the sum of low-rank widths across the shallow shared LoRA modules), and $A_t$ represents the \emph{current state} of the shared branch after task $t$ (shared parameters updated sequentially across tasks).
This concatenation aggregates low-rank directions from all shallow shared layers into a single global representation of the shared subspace.

\textbf{Basis invariance and Grassmann geometry.}
Crucially, LoRA identifies a \emph{low-rank subspace} rather than a unique parameter matrix:
any change of basis within the rank-$q$ subspace (e.g., $A\!\rightarrow\!AU$ with invertible $U$) leaves the represented subspace unchanged.
Therefore, distances in the raw parameter space are not well-defined for comparing LoRA updates across tasks.
Instead, we compare subspaces on the Grassmann manifold $\mathrm{Gr}(q,d)$, the set of all $q$-dimensional linear subspaces of $\mathbb{R}^d$, where each point corresponds to an equivalence class of bases spanning the same subspace.

We represent each subspace by its orthogonal projector (a basis-invariant representation):
\begin{equation}
P_{t-1}=A_{t-1}(A_{t-1}^\top A_{t-1})^{-1}A_{t-1}^\top,\qquad
P_t=A_t(A_t^\top A_t)^{-1}A_t^\top .
\end{equation}
The projector $P$ maps any vector to its orthogonal projection onto $\mathrm{col}(A)$, making it a canonical way to represent and compare subspaces independent of basis choices.
(Equivalently, with thin-QR $A=QR$, we compute $P=QQ^\top$, which is numerically stable and differentiable.)

We then minimize the distance between consecutive projectors,
\begin{equation}
\mathcal{L}_{\mathrm{SPP}}=\|P_t-P_{t-1}\|_F^2,
\end{equation}
which is the squared chordal distance on the Grassmann manifold and thus measures \emph{subspace} variation independent of basis choices.

Formally, for any input feature $x$,
\begin{equation}
\|P_t x - P_{t-1}x\|_2^2
\le
\|P_t-P_{t-1}\|_F^2\,\|x\|_2^2,
\label{eq:7}
\end{equation}
so reducing $\mathcal{L}_{\mathrm{SPP}}$ directly limits feature drift in the shared branch, preserving compatibility with earlier prototypes and mitigating forgetting.
Eq.~\eqref{eq:7} follows from $\|(P_t-P_{t-1})x\|_2 \le \|P_t-P_{t-1}\|_2\|x\|_2 \le \|P_t-P_{t-1}\|_F\|x\|_2$.

\subsection{Adaptive Core--Slack Alignment}

While SPP constrains the global evolution of the shared LoRA subspace, it treats all directions uniformly and does not distinguish between
(i) rotations within the existing shared subspace and
(ii) subspace expansion where previously inactive directions become activated by new tasks.
These two forms of geometric variation have different implications:
the former should be tightly regularized to preserve previously learned representations, whereas the latter reflects necessary plasticity.
To address this, we introduce Adaptive Core--Slack Alignment (ACSA), which separates and jointly controls Grassmannian core alignment and residual subspace activation.

\vspace{2pt}
\noindent\textbf{Grassmannian core alignment.}
Let the concatenated shallow shared LoRA matrices for tasks $t{-}1$ and $t$ be
$A_{t-1}\in\mathbb{R}^{d\times q_{t-1}}$ and $A_t\in\mathbb{R}^{d\times q_t}$.
We extract orthonormal bases via reduced QR decompositions:
\[
A_{t-1}=Q_{t-1}R_{t-1},\qquad A_t=Q_tR_t,
\]
where $Q_{t-1}\in\mathbb{R}^{d\times r_{t-1}}$ and $Q_t\in\mathbb{R}^{d\times r_t}$ have orthonormal columns, and $r_{t-1}=\mathrm{rank}(A_{t-1})\le q_{t-1}$, $r_t=\mathrm{rank}(A_t)\le q_t$.
The matrices $Q_{t-1}$ and $Q_t$ parameterize points on the Grassmann manifolds
$\mathrm{Gr}(r_{t-1},d)$ and $\mathrm{Gr}(r_t,d)$.

Let $r_c=\min(r_{t-1},r_t)$ and define their principal-angle cores
$\tilde Q_{t-1}=Q_{t-1}(:,1\!:\!r_c)$ and $\tilde Q_t=Q_t(:,1\!:\!r_c)$.
The quantity
\begin{equation}
s_{\mathrm{core}}^{(t)}
=
\frac{\|\tilde Q_{t-1}^\top\tilde Q_t\|_F^2}{r_c},
\label{eq:core_alignment}
\end{equation}
is a Grassmann-invariant similarity measure:
\begin{equation}
\|\tilde Q_{t-1}^\top \tilde Q_t\|_F^2
= \sum_{i=1}^{r_c} \cos^2 \theta_i,
\end{equation}
where $\theta_i$ are the principal angles between the two subspaces.
Thus,
\begin{equation}
\mathcal{L}_{\mathrm{core}}^{(t)} = 1 - s_{\mathrm{core}}^{(t)}
\end{equation}
penalizes rotations of the established shared subspace while respecting Grassmann geometry.

\vspace{2pt}
\noindent\textbf{Residual slack subspace alignment.}
Beyond the Grassmann core, new tasks may introduce additional active directions.
To quantify how these residual directions interact with previously learned subspaces,
we measure symmetric overlaps between residual bases:
\begin{equation}
\begin{aligned}
e_{\mathrm{slack}}^{(t)} =
\frac{1}{2r_c}\big(
&\mathbb{I}[r_{t-1}>r_c]\,
\|Q_t^\top Q_{t-1}(:,r_c\!+\!1{:}r_{t-1})\|_F^2 \\[-2pt]
+&\mathbb{I}[r_t>r_c]\,
\|Q_{t-1}^\top Q_t(:,r_c\!+\!1{:}r_t)\|_F^2
\big).
\end{aligned}
\label{eq:slack}
\end{equation}
Unlike Eq.~\eqref{eq:core_alignment}, these overlaps are not Grassmann invariants:
they characterize how newly activated directions interact with the established shared subspace.
When $r_{t-1}=r_t=r_c$, we define $e_{\mathrm{slack}}^{(t)}=1-s_{\mathrm{core}}^{(t)}$
to keep the regularizer active in the absence of residual dimensions.

To avoid uniformly penalizing all slack activations, we maintain an exponential moving reference:
\begin{equation}
\tau_t = \beta\tau_{t-1} + (1-\beta)e_{\mathrm{slack}}^{(t)},\qquad \beta\in(0,1),
\end{equation}
and apply an adaptive softplus penalty:
\begin{equation}
\mathcal{L}_{\mathrm{slack}}^{(t)}
=
\log\!\big(1+\exp(e_{\mathrm{slack}}^{(t)}-\tau_t)\big).
\end{equation}

Finally, ACSA interpolates between stability (core) and plasticity (slack) according to the
Grassmannian alignment level:
\begin{equation}
\mathcal{L}_{\mathrm{ACSA}}^{(t)}
=
s_{\mathrm{core}}^{(t)}\mathcal{L}_{\mathrm{core}}^{(t)}
+
(1-s_{\mathrm{core}}^{(t)})\mathcal{L}_{\mathrm{slack}}^{(t)}.
\end{equation}
When two tasks share highly similar subspaces ($s_{\mathrm{core}}^{(t)}\approx 1$),
ACSA emphasizes Grassmann core preservation.
When their alignment weakens, the emphasis shifts toward controlled plasticity in the slack directions,
allowing the shared LoRA branch to expand without catastrophic interference.

\subsection{Median-Calibrated Block Overlap}

While the shallow shared branch of Geo-LoRA maintains global geometric coherence through SPP and ACSA, the deep task-specific branch is responsible for localized semantic specialization.
In class-incremental learning, these deep LoRA modules are repeatedly deployed on the same set of Transformer blocks across tasks.
Without appropriate constraints, later tasks tend to reuse discriminative directions learned by earlier ones, resulting in subspace crowding and inter-task interference, which degrade both plasticity and retention.

To address this issue, we introduce Median-Calibrated Block Overlap (MCBO), a statistically adaptive geometric regularization that modulates representational reuse in the deep task-specific branch.
In contrast to the Grassmann-based constraints in SPP and ACSA, MCBO does not operate on a fixed-rank manifold:
the historical subspace expands as more tasks are observed, whereas each task-specific LoRA update remains low-rank.
This rank mismatch makes standard Grassmannian distances ill-suited.
Instead, MCBO measures how much the current update aligns with the accumulated historical subspace under the LoRA-induced metric, and uses a task-wise median to calibrate what constitutes ``excessive'' reuse.

Let $\mathcal{B}_{\mathrm{deep}}$ denote the set of deep Transformer blocks equipped with task-specific LoRA adapters.
For any block $b \in \mathcal{B}_{\mathrm{deep}}$, we form an orthonormal basis of the historical subspace spanned by all previous task-specific updates:
\begin{equation}
V_b \in \mathbb{R}^{d \times r_b},
\quad
V_b^{\top} V_b = I .
\label{eq:mcbo_v}
\end{equation}
The current task introduces its own LoRA update at block $b$, denoted by 
$A_b^{(t)} \in \mathbb{R}^{d \times r}$.
A natural (but scale-dependent) projection energy measuring how much $A_b^{(t)}$ lies in the historical subspace is
\begin{equation}
R_b^{(t)}
= A_b^{(t)\top} V_b V_b^{\top} A_b^{(t)} .
\label{eq:mcbo_r}
\end{equation}

Since the overall scale of $A_b^{(t)}$ may vary across tasks and blocks, we normalize this projection in the current LoRA metric by defining
\begin{equation}
G_b^{(t)}
= A_b^{(t)\top} A_b^{(t)} + \varepsilon I,
\quad
\varepsilon > 0 ,
\label{eq:mcbo_g}
\end{equation}
and the symmetrically normalized overlap matrix
\begin{equation}
\begin{aligned}
\widetilde{R}_b^{(t)}
&= G_b^{(t)-\frac{1}{2}} R_b^{(t)} G_b^{(t)-\frac{1}{2}}
= G_b^{(t)-\frac{1}{2}}
    A_b^{(t)\top} V_b V_b^{\top} A_b^{(t)}
    G_b^{(t)-\frac{1}{2}} .
\end{aligned}
\label{eq:mcbo_rtilde}
\end{equation}
This normalization removes scale dependence and isolates the geometric overlap between the current and historical low-rank subspaces under the metric induced by $A_b^{(t)}$.
In practice, task-specific LoRA updates are low-rank, so $A_b^{(t)\top} A_b^{(t)}$ can be singular or ill-conditioned; the regularization term $\varepsilon I$ ensures positive definiteness and prevents numerical failures.

The block-level overlap score is then defined as
\begin{equation}
\begin{aligned}
S_b^{(t)}
&= \frac{1}{r} \operatorname{tr}\!\big(\widetilde{R}_b^{(t)}\big)
= \frac{1}{r} \operatorname{tr}\!\Big(
A_b^{(t)\top} V_b V_b^{\top} A_b^{(t)}
\big(A_b^{(t)\top} A_b^{(t)} + \varepsilon I\big)^{-1}
\Big) .
\end{aligned}
\label{eq:mcbo_s}
\end{equation}
Intuitively, $S_b^{(t)}$ measures how strongly the current LoRA update at block $b$ reuses representational directions learned by previous tasks:
higher values indicate stronger reuse, while lower values correspond to expansion into novel task-specific directions.

Because different deep blocks naturally exhibit distinct reuse levels, MCBO avoids a global threshold and instead computes a task-wise \emph{median-calibrated reference level}:
\begin{equation}
\tau^{(t)}
= \operatorname{median}\{ S_b^{(t)} \mid b \in \mathcal{B}_{\mathrm{deep}} \} .
\label{eq:mcbo_tau}
\end{equation}
Only blocks whose overlap exceeds this typical level are softly penalized via a softplus regularization:
\begin{equation}
\mathcal{L}_{\mathrm{MCBO}}^{(t)} =
\frac{1}{|\mathcal{B}_{\mathrm{deep}}|}
\sum_{b \in \mathcal{B}_{\mathrm{deep}}}
\log\!\big(1+\exp(S_b^{(t)} - \tau^{(t)})\big).
\label{eq:mcbo_loss}
\end{equation}

\subsection{Training Objective}
Geo-LoRA integrates task learning with geometric constraints into a unified, parameter-efficient optimization framework. The overall objective is
\begin{equation}
\mathcal{L}
=
\mathcal{L}_{\mathrm{ce}}
+ \lambda_{\mathrm{spp}} \mathcal{L}_{\mathrm{SPP}}
+ \lambda_{\mathrm{acsa}} \mathcal{L}_{\mathrm{ACSA}}
+ \lambda_{\mathrm{mcbo}} \mathcal{L}_{\mathrm{MCBO}},
\end{equation}
where $\mathcal{L}_{\mathrm{ce}}$ trains the current task while the three geometric terms regulate the evolution of shared and task-specific low-rank subspaces. These constraints operate only during training and introduce no inference-time overhead.

During inference, Geo-LoRA follows the standard prototype-based protocol. The shared branch produces the backbone feature, and each task-specific branch generates its corresponding representation. Classification is performed by cosine similarity to class prototypes computed during training. This parameter-efficient scheme maintains representational consistency across tasks, enabling smooth subspace evolution and reducing catastrophic forgetting under a rehearsal-free setting.

\section{Experiments}

\subsection{Experimental Setup}

\begin{table*}[t]

\centering

\caption{Results (\%) on ImageNet-R, ImageNet-A, VTAB and CUB200 with ViT-B/16-IN1K}

\label{tab:main_results1}

\resizebox{0.97\textwidth}{!}{

\begin{tabular}{l cc cc cc cc cc cc}

\toprule

\multirow{2}{*}{Method} &

\multicolumn{2}{c}{\makecell{ImageNet-R \\ ($T{=}20$)}} &

\multicolumn{2}{c}{\makecell{ImageNet-R \\ ($T{=}10$)}} &

\multicolumn{2}{c}{\makecell{ImageNet-A \\ ($T{=}20$)}} &

\multicolumn{2}{c}{\makecell{ImageNet-A \\ ($T{=}10$)}} &

\multicolumn{2}{c}{\makecell{VTAB \\ ($T{=}5$)}} &

\multicolumn{2}{c}{\makecell{CUB-200 \\ ($T{=}10$)}} \\

& $A_T$ & $\bar{A}$ & 
$A_T$ & $\bar{A}$ & 
$A_T$ & $\bar{A}$ & 
$A_T$ & $\bar{A}$ & 
$A_T$ & $\bar{A}$ & 
$A_T$ & $\bar{A}$ \\

\midrule

L2P \cite{wang2022learning} &
68.97 & 74.16 &
71.26 & 76.13 &
38.48 & 47.16 &
42.94 & 51.40 &
80.83 & 81.19 &
65.18 & 76.12 \\

DualPrompt \cite{wang2022dualprompt} &
65.23 & 71.30 &
68.22 & 73.81 &
50.23 & 59.54 &
45.49 & 54.68 &
79.79 & 82.89 &
68.00 & 79.40 \\

CODA-Prompt \cite{smith2023coda} &
69.38 & 73.95 &
74.05 & 78.14 &
35.02 & 47.29 &
45.36 & 57.03 &
82.33 & 83.88 &
71.92 & 78.76 \\

InfLoRA \cite{liang2024inflora} &
68.89 & 76.68 &
74.75 & 80.67 &
46.21 & 59.71 &
49.20 & 60.92 &
87.63 & 88.90 &
70.82 & 81.39 \\

SD-LoRA \cite{wu2025sdlora} &
75.26 & 80.22 &
77.34 & 82.04 &
52.71 & 62.21 &
55.96 & 64.95 &
75.71 & 87.03 &
77.48 & 85.59 \\

ACMap \cite{fukuda2025adapter} &
73.75 & 79.49 &
75.06 & 80.52 &
\underline{53.29} & 63.79 &
53.21 & 64.92 &
89.35 & 92.76 &
77.62 & 85.03 \\

SEMA \cite{wang2025self} &
74.53 & 81.75 &
78.00 & 83.56 &
48.52 & 60.02 &
53.29 & 63.71 &
89.64 & 91.26 &
77.31 & 83.07 \\

BiLoRA \cite{zhu2025bilora} &
72.07 & 77.04 &
77.45 & 80.36 &
40.75 & 57.84 &
53.26 & 65.65 &
81.24 & 79.21 &
75.48 & 83.19 \\

CL-LoRA \cite{he2025cl} &
\underline{78.37} & \underline{84.39} &
\underline{80.15} & \underline{85.72} &
53.26 & \underline{64.73} &
\underline{57.43} & \underline{68.70} &
\underline{93.06} & \underline{94.05} &
\underline{78.12} & \underline{86.32} \\

\midrule

Geo-LoRA (Ours) &

\textbf{79.62} & \textbf{84.93} &

\textbf{81.21} & \textbf{86.46} &

\textbf{54.58} & \textbf{65.22} &

\textbf{59.62} & \textbf{69.90} &

\textbf{93.96} & \textbf{94.71} &

\textbf{78.63} & \textbf{88.35} \\

\bottomrule

\end{tabular}

}

\vspace{-3mm}

\end{table*}

\begin{table*}[t]

\centering

\caption{Results (\%) on ImageNet-R, ImageNet-A, VTAB and CUB200 with ViT-B/16-IN21K}

\label{tab:main_results}

\resizebox{0.97\textwidth}{!}{

\begin{tabular}{l cc cc cc cc cc cc}

\toprule

\multirow{2}{*}{Method} &

\multicolumn{2}{c}{\makecell{ImageNet-R \\ ($T{=}20$)}} &

\multicolumn{2}{c}{\makecell{ImageNet-R \\ ($T{=}10$)}} &

\multicolumn{2}{c}{\makecell{ImageNet-A \\ ($T{=}20$)}} &

\multicolumn{2}{c}{\makecell{ImageNet-A \\ ($T{=}10$)}} &

\multicolumn{2}{c}{\makecell{VTAB \\ ($T{=}5$)}} &

\multicolumn{2}{c}{\makecell{CUB-200 \\ ($T{=}10$)}} \\

& $A_T$ & $\bar{A}$ &
$A_T$ & $\bar{A}$ &
$A_T$ & $\bar{A}$ &
$A_T$ & $\bar{A}$ &
$A_T$ & $\bar{A}$ &
$A_T$ & $\bar{A}$ \\

\midrule

L2P \cite{wang2022learning} &

62.15 & 68.35 &
64.94 & 70.33 &
39.30 & 46.67 &
37.62 & 39.81 &
76.41 & 78.96 &
63.24 & 75.01 \\

DualPrompt \cite{wang2022dualprompt} &

66.89 & 73.07 &
62.24 & 74.63 &
48.78 & 58.45 &
47.45 & 56.43 &
80.94 & 82.51 &
69.22 & 79.80 \\

CODA-Prompt \cite{smith2023coda} &

67.53 & 73.64 &
72.15 & 77.51 &
37.06 & 50.73 &
51.61 & 60.70 &
89.49 & 92.27 &
70.11 & 80.74 \\

InfLoRA \cite{liang2024inflora} &

71.01 & 77.28 &
75.65 & 80.82 &
41.61 & 56.84 &
47.04 & 56.91 &
87.16 & 88.83 &
70.47 & 81.02 \\

SD-LoRA \cite{wu2025sdlora} &

74.31 & 80.00 &
77.01 & 81.92 &
45.89 & 56.60 &
50.10 & 59.17 &
72.77 & 85.24 &
63.74 & 82.61 \\

ACMap \cite{fukuda2025adapter} &

72.95 & 78.82 &
73.50 & 79.50 &
\underline{51.58} & 62.56 &
56.19 & 65.19 &
87.56 & 91.21 &
76.42 & 83.09 \\

SEMA \cite{wang2025self} &

69.60 & 77.84 &
74.82 & 81.39 &
46.94 & 58.36 &
45.95 & 57.89 &
90.86 & 91.99 &
\underline{79.62} & 84.95 \\

BiLoRA \cite{zhu2025bilora} &

72.41 & 79.28 &
77.95 & 81.52 &
40.75 & 55.18 &
48.91 & 61.47 &
81.20 & 80.47 &
78.34 & 85.04 \\

CL-LoRA \cite{he2025cl} &

\underline{77.42} & \underline{83.57} &

\underline{80.20} & \underline{85.41} &

51.35 & \underline{63.75} &

\underline{58.09} & \underline{69.30} &

\underline{93.31} & \underline{93.51} &

79.32 & \underline{88.12} \\

\midrule

Geo-LoRA (Ours) &

\textbf{78.30} & \textbf{84.12} &

\textbf{80.95} & \textbf{86.15} &

\textbf{52.73} & \textbf{64.85} &

\textbf{58.79} & \textbf{69.53} &

\textbf{93.72} & \textbf{94.85} &

\textbf{80.41} & \textbf{89.05} \\

\bottomrule

\end{tabular}

}

\vspace{-3mm}

\end{table*}

\textbf{Datasets.}
We evaluate Geo-LoRA on six rehearsal-free CIL benchmarks covering natural images, distribution shifts, fine-grained recognition, and heterogeneous visual domains.
CIFAR-100 contains 100 natural categories.
ImageNet-R and ImageNet-A each include 200 ImageNet-derived classes with artistic or adversarial distribution shifts.
CUB200 comprises 200 fine-grained bird species.
Following prior work, we construct a 50-class VTAB subset spanning diverse visual domains.
OmniBenchmark provides 300 heterogeneous classes from multiple visual sources.
All datasets are split into $T$ equal-sized tasks to create CIL sequences of varying lengths.

\textbf{Compared Methods.}
We compare Geo-LoRA with representative rehearsal-free PEFT-based CIL methods across prompt tuning, lightweight adapters, and LoRA variants, including L2P, DualPrompt, CODA-Prompt, InfLoRA, SD-LoRA, ACMap, SEMA, BiLoRA, and CL-LoRA.
This suite covers the major parameter-efficient paradigms and forms a comprehensive evaluation.

\textbf{Evaluation Metrics.}
Following standard protocol, we report the average accuracy
where $A_t$ is accuracy over all seen classes after task $t$,
and the final accuracy $A_T$ after completing the last task.
These metrics assess both overall performance and long-term retention.

\textbf{Implementation Details.}
We implement all models in PyTorch\cite{paszke2019pytorch} and PILOT\cite{sun2025pilot}, using the same ViT-B/16\cite{dosovitskiy2020image} backbone with N = 12 Transformer blocks. Following prior CIL work, we adopt two representative pre-trained models—ViT-B/16-IN1K and ViT-B/16-IN21K—both initialized from ImageNet-21K, with the former further fine-tuned on ImageNet-1K.
For LoRA, we use rank $r = 10$ with $A \in \mathbb{R}^{768 \times 10}$ and $B \in \mathbb{R}^{10 \times 768}$.
Task-shared LoRA adapters are inserted into the first l = 6 blocks, and task-specific adapters into the remaining 6 blocks.
All baseline implementations are based on publicly released official code or reported numbers from published papers.
Except for SD-LoRA, which requires substantially more GPU memory and is therefore run on an NVIDIA H800, all other methods—including our Geo-LoRA—are trained and evaluated on NVIDIA RTX 3090.

\subsection{Experimental Results}
\begin{figure*}[!ht]
    \begin{center}
	\includegraphics[width=1.0\linewidth]{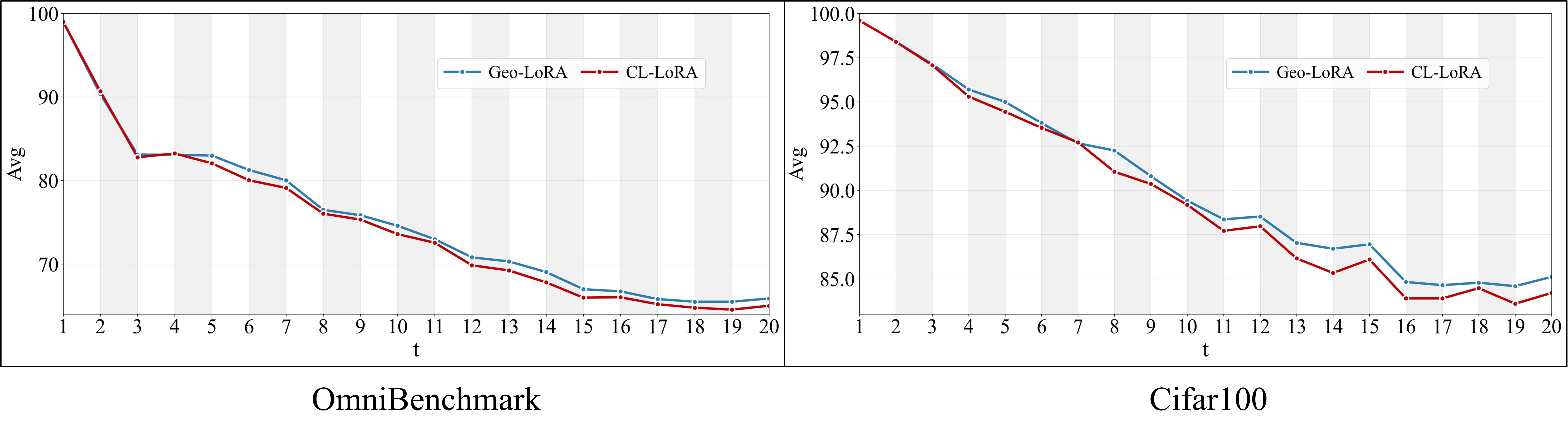}
	\caption{Results (\%) on OmniBenchmark and Cifar100
across training stages with ViT-B/16-IN1K ($T{=}20$).}
	\label{fig:framework}
    \end{center}    
\end{figure*}

As shown in Tables~\ref{tab:main_results1} and~\ref{tab:main_results}, Geo-LoRA consistently surpasses all baselines across varying task numbers and both 1k- and 21k-pretrained ViT backbones. The improvements are especially notable on challenging benchmarks such as ImageNet-R, ImageNet-A, VTAB, and CUB200.
On ImageNet-R/A with 10-task and 20-task sequences—where artistic transformations and adversarial perturbations induce strong distribution shifts—Geo-LoRA achieves the best accuracy on both PTMs, for instance, our method achieves accuracy improvements of 2.19\% and 1.25\% on ImageNet-A (10 tasks) and ImageNet-R (20 tasks), respectively, when using the 1K pretrained model.
Figure~\ref{fig:framework} further compares Geo-LoRA with the latest CL-LoRA on OmniBenchmark and CIFAR-100 (1k PTM). Geo-LoRA achieves higher final accuracy and maintains an advantage across most tasks, demonstrating more favorable stability–plasticity dynamics.
To assess generalization under diverse regimes, we additionally run 5-task short-horizon and 40-task long-horizon experiments on ImageNet-A/R using the 1k backbone. As shown in Tables~\ref{tab:40task_results} and~\ref{tab:5task_results}, Geo-LoRA remains the top performer even in these extreme settings, confirming strong scalability.

Meanwhile, we further evaluate the computational overhead on ImageNet-A with the 21k-pretrained backbone under both 10-task and 20-task settings. As reported in Table~\ref{tab:time}, we measure the total training time as well as the inference time during the final evaluation. Compared with the baselines, Geo-LoRA incurs only a slight increase in overall training time, while its inference time remains nearly unchanged. Given the clear performance gains, we believe this modest training overhead is well justified.

\begin{table}[t]
\centering
\caption{Results (\%) on ImageNet-R and ImageNet-A with ViT-B/16-IN1K ( $T{=}40$).}
\label{tab:40task_results}
\renewcommand{\arraystretch}{1}
\begin{tabular}{l cc cc}
\toprule
\multirow{2}{*}{Method} &
\multicolumn{2}{c}{\makecell{ImageNet-R}} &
\multicolumn{2}{c}{\makecell{ImageNet-A}} \\
& $A_T$ & $\bar{A}$ & $A_T$ & $\bar{A}$ \\
\midrule
L2P        & 60.62 & 65.82 & 27.42 & 38.09 \\
DualPrompt & 61.73 & 67.41 & 42.75 & 52.26 \\
CODA-Prompt & 63.93 & 70.39 & 27.93 & 38.39 \\
InfLoRA    & 64.51 & 73.22 & 40.02 & 52.97 \\
SD-LoRA    & 70.07 & 76.21 & 42.49 & 53.07 \\
ACMap      & 71.10 & 77.78 & \underline{43.55} & 53.75 \\
SEMA       & 65.38 & 74.89 & 32.98 & 50.44 \\
BiLoRA     & 64.05 & 68.99 & 31.80 & 47.26 \\
CL-LoRA    & \underline{73.37} & \underline{81.54} & 43.45 & \underline{56.97} \\
\midrule
Geo-LoRA (Ours) & \textbf{73.90} & \textbf{82.19} & \textbf{44.04} & \textbf{57.53} \\
\bottomrule
\end{tabular}
\end{table}

\begin{table}[t]
\centering
\caption{Results (\%) on ImageNet-R and ImageNet-A with ViT-B/16-IN1K ($T{=}5$).}
\label{tab:5task_results}
\renewcommand{\arraystretch}{1}
\begin{tabular}{l cc cc}
\toprule
\multirow{2}{*}{Method} &
\multicolumn{2}{c}{\makecell{ImageNet-R }} &
\multicolumn{2}{c}{\makecell{ImageNet-A }} \\
& $A_T$ & $\bar{A}$ & $A_T$ & $\bar{A}$ \\
\midrule
L2P        & 73.04 & 76.94 & 40.21 & 49.02 \\
DualPrompt & 69.99 & 72.24 & 54.72 & 62.93 \\
CODA-Prompt & 76.63 & 80.30 & 42.38 & 50.21 \\
InfLoRA    & 76.95 & 81.81 & 49.37 & 63.27 \\
SD-LoRA    & 77.30 & 82.30 & 55.83 & 65.15 \\
ACMap      & 76.09 & 80.02 & 63.38 & 72.14 \\
SEMA       & 75.92 & 82.27 & 56.47 & 64.62 \\
BiLoRA     & 75.13 & 79.33 & 43.27 & 59.21 \\
CL-LoRA    & \underline{82.05} & \underline{86.15} & \underline{63.39} & \underline{72.32} \\
\midrule
Geo-LoRA (Ours) & \textbf{82.48} & \textbf{86.85} & \textbf{64.25} & \textbf{72.83} \\
\bottomrule
\end{tabular}
\end{table}

\begin{figure*}[!ht]
    \begin{center}
	\includegraphics[width=1.0\linewidth]{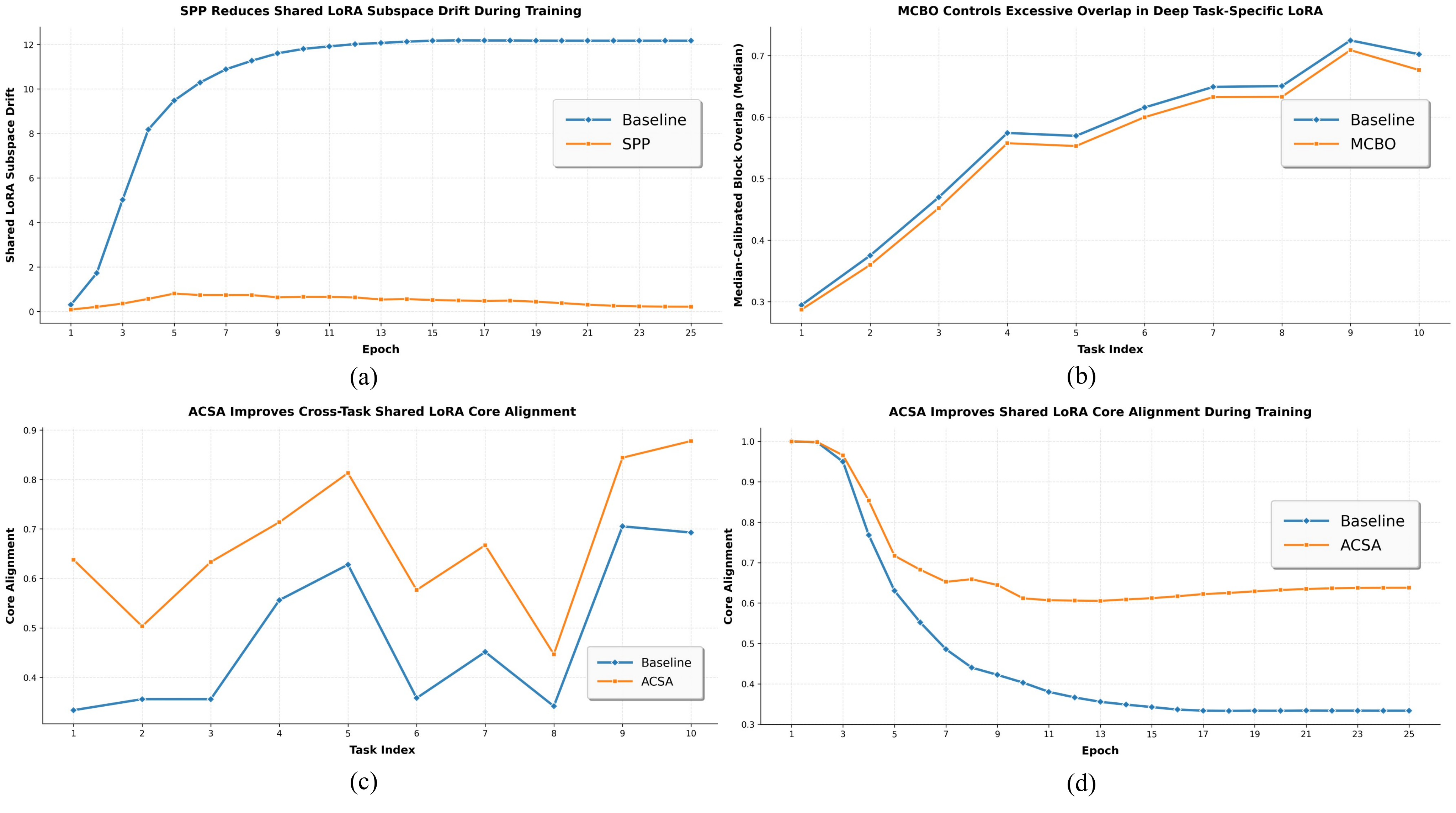}
	\caption{Empirical validation of the geometric mechanisms in Geo-LoRA.
(a) SPP suppresses shared subspace drift over training epochs.
(b) MCBO reduces median-calibrated overlap in deep task-specific LoRA across tasks, indicating controlled reuse of historical directions.
(c) ACSA improves cross-task core alignment of shared subspaces.
(d) ACSA maintains higher core alignment during training.
These results demonstrate that the proposed constraints explicitly regulate subspace drift, overlap, and alignment, jointly enabling controlled geometric evolution of LoRA subspaces.
}
	\label{fig:exp}
    \end{center}
\end{figure*}

\subsection{Experimental Analysis}
The superior performance of Geo-LoRA on both short and long CIL sequences can be attributed to its explicit regulation of the evolution of feature subspaces—keeping representations geometrically aligned across tasks—whereas prior PEFT methods mainly act at token level and may not directly constrain the cumulative drift that can destabilize decision boundaries.

Relative to single-branch LoRA approaches (e.g., BiLoRA and SD-LoRA), Geo-LoRA exhibits more reliable behavior by mitigating deep-layer subspace crowding. Single-branch LoRA often repeatedly reuses a narrow set of discriminative directions, limiting the capacity to encode genuinely novel task-specific variations. In contrast, SPP and ACSA stabilize the shared branch to accumulate task-agnostic structure coherently, while MCBO encourages the deep branch to explore new, non-conflicting directions. Consequently, Geo-LoRA better preserves earlier feature semantics while learning richer and less entangled representations for new tasks, improving stability and plasticity under comparable parameter budgets.

Compared with prior dual-LoRA designs such as CL-LoRA, Geo-LoRA further benefits from a unified projection-based geometric principle that explicitly coordinates the evolution of shared and task-specific subspaces. This coordination enables the shared branch to absorb cross-task common structure in a geometry-consistent manner, while the task-specific branch focuses on localized discrimination without collapsing into previously occupied directions, reducing interference within fine-grained subspaces.

Overall, these results suggest that geometry-aware subspace regulation is a principled and practical mechanism for parameter-efficient continual learning, improving long-term retention and task-specific adaptation across diverse and distribution-shifted CIL benchmarks.

\begin{table}[t]
\centering
\caption{Total training time (T) and inference time (I) comparison on ImageNet-A.}
\begin{tabular}{lcccc}
\toprule
Method & T(20task) & I(20task) & T(10task) & I(10task) \\
\midrule
Baseline & 44.37 min & 77.84 s & 30.83 min & 39.61 s \\
Geo-LoRA & 46.85 min & 77.60 s & 32.27 min & 39.54 s \\
\bottomrule
\end{tabular}

\label{tab:time}
\end{table}



\begin{table}[t]
\centering
\caption{Ablation of geometric objectives on the ImageNet-A dataset with a 1K-pretrained model.}
\begin{tabular}{lccc}
\toprule
Method & 20task & 10task & 5task  \\
\midrule
Baseline & 53.26 & 57.43 & 63.39 \\
+ SPP + ACSA & 53.84  & 58.42& 63.71 \\
+ SPP + MCBO & 54.10 & 59.04 & 63.98  \\
+ ACSA + MCBO & 53.92 & 58.97& 63.84\\
Full Geo-LoRA & \textbf{54.58} & \textbf{59.62} & \textbf{64.25}  \\
\bottomrule
\end{tabular}

\label{tab:mod}
\end{table}

\subsection{Representation Dynamics Analysis}
Beyond performance metrics, we further analyze the geometric dynamics targeted by Geo-LoRA from three aspects: (i) subspace drift in the shallow shared branch, (ii) cross-task core alignment of shared subspaces, and (iii) excessive reuse in deep task-specific blocks. As shown in Fig.~\ref{fig:exp}, SPP markedly suppresses shared-subspace drift during training, ACSA consistently improves and maintains higher core alignment across tasks and epochs, and MCBO reduces median-calibrated overlap in deep task-specific LoRA blocks, indicating more controlled reuse of historical directions. These results provide direct empirical evidence that Geo-LoRA explicitly regulates drift, alignment, and overlap during continual adaptation.

\subsection{Ablation Study}
We conduct ablation studies to evaluate the contribution of each component and the method design in Geo-LoRA.

\textbf{Geometric objectives.}
As shown in Table~\ref{tab:mod}, all three components provide substantial and complementary benefits: SPP improves cross-task consistency, ACSA refines the stability–plasticity balance, and MCBO strengthens deep-layer specialization. Removing any single term leads to a clear drop in performance, highlighting that Geo-LoRA relies on their joint interaction rather than any individual objective.

\textbf{SPP normalization.}
Table~\ref{tab:No} compares the Grassmann projector $A_t (A_t^\top A_t)^{-1} A_t^\top$ with the covariance-based normalization $A_{t-1}^\top A_{t-1} + \varepsilon I$. The projector consistently yields higher accuracy, demonstrating superior geometric stability and robustness under low-rank updates.

\begin{table}[h]
\centering
\renewcommand{\arraystretch}{1.25}
\setlength{\tabcolsep}{6pt}
\caption{Comparison of SPP normalization schemes on ImageNet-R (20 tasks).}
\begin{tabular}{c|c c}
\hline
Normalization Choice & $A_T$ (\%) & $\bar{A}$ (\%)\\
\hline
$A_{t-1}(A_{t-1}^{\top}A_{t-1}+\varepsilon I)^{-1}A_{t-1}^{\top}$ & 79.27 &84.77\\
$A_{t-1}(A_{t-1}^{\top}A_{t-1})^{-1}A_{t-1}^{\top}$ & \textbf{79.62} &\textbf{84.93} \\
\hline
\end{tabular}

\label{tab:No}
\end{table}



\section{Conclusion}
We introduced Geo-LoRA, a unified geometric framework for continual low-rank adaptation that reframes LoRA updates as the evolution of task-dependent subspaces. By regulating the geometry of both shared and task-specific low-rank structures, Geo-LoRA provides a principled way to control how representational subspaces drift, expand, and interact across tasks.
Through the complementary roles of SPP, ACSA, and MCBO, Geo-LoRA establishes a coherent hierarchy in which shallow layers preserve global consistency while deep layers manage localized plasticity. This geometry-aware design delivers strong accuracy and forgetting reductions over prior adapter-based CIL methods, without additional model capacity or replay.
More broadly, our formulation suggests that continual adapter tuning can be viewed as managing subspace trajectories on evolving low-rank manifolds. We hope this perspective encourages future work on geometry-aware adaptation and scalable continual tuning for vision and multimodal foundation models.


\bibliographystyle{ACM-Reference-Format}
\bibliography{sample-base}

\end{document}